%% file: tmlr.tex
\documentclass[10pt]{article} % For LaTeX2e
\usepackage[accepted]{tmlr}
\input{math_commands.tex}

\usepackage{hyperref}
\usepackage{amsmath}
\usepackage{amssymb}
\usepackage{mathtools}
\usepackage{amsthm}
\usepackage{adjustbox}
\usepackage{enumitem}
\usepackage[capitalize,noabbrev]{cleveref}
\usepackage{microtype}
\usepackage{graphicx}
\usepackage{subfigure}
\usepackage{booktabs} % for professional tables
\usepackage{bbm}
\usepackage[utf8]{inputenc} % allow utf-8 input
\usepackage[T1]{fontenc}    % use 8-bit T1 fonts
\usepackage{url}            % simple URL typesetting
\usepackage{booktabs}       % professional-quality tables
\usepackage{amsfonts}       % blackboard math symbols
\usepackage{nicefrac}       % compact symbols for 1/2, etc.
\usepackage{microtype}      % microtypography
\usepackage{xcolor}         % colors
\usepackage{pifont}
\newcommand{\cmark}{\text{\ding{51}}}
\newcommand{\xmark}{\text{\ding{55}}}
\usepackage{url}
\usepackage{wrapfig}
\usepackage[framemethod=TikZ]{mdframed}
\mdfsetup{%
    middlelinecolor =   none,
    middlelinewidth =   1pt,
    backgroundcolor =   blue!5,
    roundcorner     =   5pt,
}
\newmdenv[
    middlelinecolor=none,
    middlelinewidth=1pt,
    backgroundcolor=blue!5,
    roundcorner=5pt
]{bluebox}

\newmdenv[
    middlelinecolor=none,
    middlelinewidth=1pt,
    backgroundcolor=gray!20,
    roundcorner=5pt
]{graybox}
\usepackage{xcolor}
\usepackage{tcolorbox}
\newtcbox{\grayboxtext}[1][]{
    on line,
    colframe=gray!20,
    colback=gray!20,
    boxrule=0.5pt,
    arc=4pt,
    boxsep=0pt,
    left=2pt,
    right=2pt,
    top=2pt,
    bottom=2pt,
    #1
}
\theoremstyle{plain}
\newtheorem{theorem}{Theorem}[section]

\theoremstyle{definition}

\theoremstyle{remark}
\newtheorem{remark}[theorem]{Remark}

\title{Inverse Reinforcement Learning Meets Large Language Model Post-Training: Basics, Advances, and Opportunities}

\author{\name Hao Sun \email hs789@cam.ac.uk \\
      \addr Department of Applied Mathematics and Theoretical Physics\\
      University of Cambridge \\
      Cambridge, United Kingdom
      \AND
      \name Mihaela van der Schaar \email mv472@cam.ac.uk \\
      \addr Department of Applied Mathematics and Theoretical Physics\\
      University of Cambridge \\
      Cambridge, United Kingdom}

\begin{document}

\maketitle

\begin{abstract}
In the era of Large Language Models (LLMs), alignment has emerged as a fundamental yet challenging problem in the pursuit of more reliable, controllable, and capable machine intelligence. The recent success of reasoning models and conversational AI systems has underscored the critical role of reinforcement learning (RL) in enhancing these systems, driving increased research interest at the intersection of RL and LLM alignment.
This paper provides a comprehensive review of recent advances in LLM alignment through the lens of inverse reinforcement learning (IRL), emphasizing the distinctions between RL techniques employed in LLM alignment and those in conventional RL tasks. In particular, we highlight the necessity of constructing neural reward models from human data and discuss the formal and practical implications of this paradigm shift.
We begin by introducing fundamental concepts in RL to provide a foundation for readers unfamiliar with the field. We then examine recent advances in this research agenda, discussing key challenges and opportunities in conducting IRL for LLM alignment. Beyond methodological considerations, we explore practical aspects, including datasets, benchmarks, evaluation metrics, infrastructure, and computationally efficient training and inference techniques.
Finally, we draw insights from the literature on sparse-reward RL to identify open questions and potential research directions. By synthesizing findings from diverse studies, we aim to provide a structured and critical overview of the field, highlight unresolved challenges, and outline promising future directions for improving LLM alignment through RL and IRL techniques.
\end{abstract}

\section{Motivation: Reinforcement Learning in the Era of Large Language Models}
\subsection{The Success of Large-Scale Data-Driven Models}
In the era of large foundation models, great success has been achieved by scaling up training compute, data, and the number of model parameters~\citep{vaswani2017attention,kaplan2020scaling,hoffmann2022training,zhang2024when}. And such great success spans in many fields from natural language generation~\citep{achiam2023gpt,meta2024introducing,team2024gemini}, understanding~\citep{devlin2019bert,liu2019roberta,raffel2020exploring}, high-resolution image generation~\citep{ramesh2021zero,ramesh2022hierarchical,podell2023sdxl,zhang2023adding}, editing~\citep{hertz2022prompt,zhang2025scaling}, audio~\citep{kong2020diffwave,copet2023simple,wang2023neural} and video generation~\citep{brooks2024video}, decision-making and control~\citep{reed2022generalist,brohan2023rt,bousmalis2023robocat,driess2023palm}.

Among those large-scale, successful data-driven models, we are particularly interested in the Large Language Models (LLMs), given their high potential of transparency through natural language~\citep{liao2023ai,lindsey2025biology}, the recent progress of applying those models in general-purpose assistant systems~\citep{ouyang2022training}, and agentic use-cases to perform deep analysis~\citep{openai2025deepresearch}. 

However, while LLMs can understand and follow users' instructions~\citep{zhou2023instruction}, quickly adapt to new tasks~\citep{brown2020language}, and can have reasoning abilities to finish complex tasks~\citep{wei2022chain,kojima2022large,guo2025deepseek}, those systems can not always do self-correction by themselves~\citep{huang2023large,kamoi2024can} and keep the system continue improving.

\subsection{The Success of Large Scale Reinforcement Learning}
Since the success of Reinforcement Learning (RL) for Atari games and AlphaGo~\citep{mnih2013playing,silver2016mastering}, the ability of RL in achieving super-human performance has been demonstrated in board games~\citep{silver2017mastering,schrittwieser2020mastering}, real-time strategy games~\citep{vinyals2019grandmaster,berner2019dota}, and many other applications ranging from chip design to algorithmic optimization~\cite{mirhoseini2021graph,fawzi2022discovering,mankowitz2023faster}. By interacting with the environment, those RL systems can keep improving their abilities to solve the training tasks and finally achieve super-human performance. 

While RL can achieve super-human performance and create novel solutions to problems, the transparency of RL systems remains a non-trivial challenge~\citep{qing2022survey,milani2022survey}. It's challenging for humans to identify, understand, and learn from those creative behaviors~\citep{menick2016move,bory2019deep,zahavy2023diversifying}. 

\subsection{Combining the Success from Both Sides: RL Meets LLM Post-Training}
Given the success of RL and LLM in their respective domains, combining the success from both sides becomes promising. From an RL-centered perspective, if we can harness LLMs to achieve superhuman performance, natural language may serve as the ideal interface to leverage RL's creativity to inspire humans; from the LLM-centered standpoint, RL can grant LLMs the ability to continually enhance performance on reward-defined tasks. 
\paragraph{LLM Alignment and Post-Training}
In this paper, we exchangably use \textit{alignment} and \textit{post-training} to denote optimizing pre-trained LLMs aimed at gaining specific capabilities. RL naturally aligns with such a learning paradigm as well as the capability can be quantified as a reward.
\paragraph{RL in Conversational AI}
In general-purpose dialogue systems, RLHF is proven to be an effective approach to enhance LLMs' abilities through preference annotation, and this is especially useful in tasks where golden evaluation metrics are difficult or impossible to define~\citep{christiano2017deep,bai2022constitutional,stiennon2020learning,bai2022training}. Further investigation on alternative approaches explored different aspects of improving such a paradigm~\citep{rafailov2023direct,ethayarajh2024kto,zhao2023slic,liu2023statistical,ji2024towards,meng2024simpo,yin2024relative,sun2024generalizing,azar2024general}. The enormous user base and their feedback provide
OpenAI with a continuous stream of data to model user preferences and enhance the experience. 

\paragraph{RL in Mathematical Reasoning}
In mathematics, AlphaProof and AlphaGeometry2 won silver medals at the International Mathematical Olympiad (IMO)~\citep{deepmind2024imo,trinh2024solving}. Moreover, DeepSeek-R1~\citep{guo2025deepseek} demonstrated the power of RL in mathematical reasoning and more general reasoning tasks. Through the technique of RL, LLMs can learn the behavior of deep thinking or self-reflection, and then improve their ability in solving tasks by generating more tokens~\citep{xu2025towards}.

\paragraph{Opportunities and Key Challenges}
What are the key challenges in scaling up RL to a wider range of LLM tasks and applications? 
First, lacking reward signals. In most tasks, we do not have rule-based reward signals as in the math or coding tasks. In those cases, efficient reward modeling becomes vitally important, and this will be the focus of the 3rd section of this paper. 
Second, the demand for computing. The prohibitively high cost in compute hinders the open-source development of this field. To help alleviating such a challenge, we will introduce a reward model infrastructure to conduct Inverse RL research on LLM alignment efficiently. With such an infrastructure, researchers without GPUs can also efficiently verify their ideas. For the 3rd challenge, although we have lots of algorithms in RL, there is no silver bullet. We need to consider the properties of different LLM alignment tasks. Hence, we are motivated to have this paper, which tries to bridge the gap between Inverse RL and LLM alignment could be helpful for potential future research.

\section{Revisiting the Foundations of Reinforcement Learning under an LLM Context}
In reinforcement learning, an agent interacts with the external environment to collect feedback and observations. The objective of such a learning process is to maximize the long-term return~\citep{sutton1998reinforcement}. 
\subsection{Markov Decision Processes}
In Markov Decision Processes, decisions are made in discrete time steps and affect the state of the environment in the subsequent step.
Formally, an MDP is denoted as $\mathcal{M} = \{\mathcal{S},\mathcal{A},\mathcal{T},\mathcal{R},\rho_0,\gamma\}$, where $\mathcal{S}\subset \mathbb{R}^{d}$ denotes the $d$-dim state space, $\mathcal{A}$ is the action space. Broadly, the environment includes $\mathcal{T}$ and $\mathcal{R}$, the former denotes the transition dynamics $\mathcal{T}: \mathcal{S}\times \mathcal{A} \mapsto \Delta(\mathcal{S})$ that controls transitions between states, and the reward function $\mathcal{R}:\mathcal{S}\times\mathcal{A}\mapsto \mathbb{R}$ provides feedback. $\rho_0 = p(s_0)\in\Delta(\mathcal{S})$ denotes the initial state distribution. $\gamma$ is the discount factor that trades off between short-term and long-term returns.

To solve an MDP problem, the high-level idea is fairly simple --- the agent should learn to \textit{discover} and \textit{repeat} successful actions and trajectories. Formally in literature, the \textit{discover} is referred to as \textit{exploration}, and \textit{repeat} is \textit{exploitation}~\citep{sutton1998reinforcement}.

Although the idea is simple and elegant, the practice and implementation are far from trivial. One fact in the RL literature is that, while some RL algorithms can be better than others in some tasks, there is no single RL algorithm that performs best on every task. Each algorithm comes with its own assumptions, strengths, and limitations. The choice of algorithm should be determined by environmental properties, and sometimes resource constraints. Table~\ref{tab:1} shows some examples of different tasks and corresponding successful algorithms, categorized by the structure of their action space $\mathcal{A}$, state space $\mathcal{S}$, reward signal $\mathcal{R}$, algorithmic approaches, and whether their transition dynamics $\mathcal{T}$ are known.

\begin{table}[h]
\centering
\caption{\small Representative RL Tasks and Characteristics}
\begin{adjustbox}{max width=\textwidth}
\begin{tabular}{lllllll}
\toprule
\textbf{Task} & \textbf{$\mathcal{A}$} & \textbf{$\mathcal{S}$} & \textbf{$\mathcal{R}$} & \textbf{Algorithms} & \textbf{$\mathcal{T}$} \\
\midrule
Atari-Dense & Disc. $\sim 10^1$ & Image & Dense & DQN & Unknown \\
Atari-Explore & Disc. $\sim 10^1$ & Image & Sparse & Curiosity-Driven & Unknown \\
Board Game & Disc. $\sim 10^2$ & Disc. $\sim 10^{100}$ & Sparse & MCTS, Self-Play & Known \\
Dota2 & Disc. $\sim 10^6$ & PO, Mixed & Mixed & (MA)PPO & Unknown \\
StarCraft & Disc. $\sim 10^{26}$ &  PO, Mixed & Mixed & BC, AC, League & Unknown \\
Robotics-GC & Cont. Dim $\sim 10^2$ & Cont. Dim $\sim 10^2$ & Sparse & Hindsight Exp. Replay & Unknown \\
Locomotion & Cont. Dim $\sim 10^2$ & Cont. Dim $\sim 10^2$ & Dense & SAC, TD3, TD7 & Unknown \\
Reasoning & Disc. $\sim 10^6$ & $\mathcal{V}^C$ & Sparse & GRPO & Known \\
RLHF & Disc. $\sim 10^6$ & $\mathcal{V}^C$ & Noisy Sparse & PPO, DPO, REINFORCE & Known \\
\bottomrule
\end{tabular}
\label{tab:1}
\end{adjustbox}
\end{table}

In Atari games, pioneered by the DQN model \cite{mnih2015human}, the agent operates in a discrete action space with visual input and mostly dense rewards. However, when modified for sparse-reward settings (e.g., exploration-focused variants), intrinsic motivation and curiosity-driven approaches become necessary \cite{pathak2017curiosity}.
Board games like Go involve very large discrete state spaces and sparse rewards, where planning and search-based methods such as MCTS and self-play have proven highly effective \cite{silver2017mastering}.
Dota 2 and StarCraft exemplify complex, partially observable (PO) hybrid state spaces, multi-agent environments with both sparse and dense reward components. These settings have motivated scalable and distributed algorithms like (Multi-Agent) PPO \cite{berner2019dota} and league-based training with off-policy actor-critic methods \cite{vinyals2019grandmaster}.
Multi-goal robotic manipulation tasks typically involve continuous state and action spaces, and sparse goal-conditioned rewards. Hindsight Experience Replay (HER) has shown promise in addressing the challenge of learning from failures in such settings \cite{andrychowicz2017hindsight}. In locomotion tasks, methods such as TD3, SAC, and TD7~\citep{fujimoto2018addressing,haarnoja2018soft,fujimoto2023sale} are widely used.
LLM-based reasoning tasks operate over a large discrete action space (e.g., vocabulary tokens raised to the length of context length) and often rely on sparse or delayed rewards. Recently, GRPO \cite{deepseek2024grpo} has been proposed to handle these cases more effectively.
Finally, Reinforcement Learning from Human Feedback (RLHF) tasks focus on aligning language models with human preferences. These involve noisy, implicit reward signals derived from pairwise or ranked feedback. Algorithms such as PPO, DPO, and REINFORCE are widely used in this domain \cite{christiano2017deep, rafailov2023direct, williams1992simple}.

\begin{mdframed}[innertopmargin=0pt,leftmargin=0pt, rightmargin=0pt, innerleftmargin=10pt, innerrightmargin=10pt, skipbelow=0pt]
\textbf{\textcolor{brown}{Take-away} There is no silver bullet in RL.} The choice of algorithm should be determined by environmental properties (state space, action space, transition dynamics, reward sparsity etc.), and resource constraints.
\end{mdframed}

\subsection{Characterizing LLM Generation in an MDP Framework: The Challenge of Missing Reward}
Using the MDP framework discussed above, we can formally describe the LLM token generation process. Let $C$ denote the context window size and $\mathcal{V}$ denote the vocabulary, including the special tokens like \texttt{[EOS]} and \texttt{[MASK]}. The MDP is instantiated as follows: 
State space $\mathcal{S} = \mathcal{V}^C$; action space $\mathcal{A}=\mathcal{V}$; transition dynamics is \textbf{deterministic and known}: $s' = \mathcal{T}(s,a) = \texttt{Concat}(s,a) = [s, a] $; We consider states containing an \texttt{[EOS]} token as absorbing states, meaning $ \forall a: s' = \mathcal{T}(s,a) = s ~\textrm{if}~ \texttt{[EOS]}\in s$; 
an LLM $\ell$, serving as policy $\pi = \ell$, generates the next token $a\in\mathcal{A}$ based on the current context $s\in\mathcal{S}$; The initial state distribution of queries is $\rho_0$, and $T$ represents the maximal number of new tokens in a generation. i.e., $T$ is the maximal number of transitions in the MDP.
For instance, in the following case, the context window length $C\ge7$ and $T=2$, an initial state $s_0\sim\rho_0$, sampled from the initial prompt or user query distribution $\rho_0$, is given as follows:
\begin{equation*}
    s_0 = \big[\texttt{ The | color | of | 
 the | sky |\hspace{1pt}[MASK]\hspace{1pt}|\hspace{1pt}[MASK]}\big],
\end{equation*}
when the language model policy $\pi$ selects a new token ``$\texttt{is}$'' from the vocabulary $\mathcal{V}$, the next state deterministically becomes
\begin{equation*}
\begin{split}
    s_1 = \texttt{Concate}(s_0, a_0=\texttt{is})= \big[\texttt{ The | color | of | 
 the | sky | is |\hspace{1pt}[MASK]}\big],
\end{split}
\end{equation*}
the generation process continues until either the \texttt{[EOS]} token is selected, the maximal context window size is reached, or the maximal decision steps $T$ is reached. In this example, the final generated context could be:
\begin{equation*}
\begin{split}
    s_2 = \texttt{Concate}(s_1, a_1=\texttt{blue})= \big[\texttt{ The | color | of | 
 the | sky | is | blue }\big].
\end{split}
\end{equation*}

When it comes to the reward function $\mathcal{R}$, its definition is less clear and non-trivial.
In LLM generation, there is no external reward verifier, such as ``winning a game'' or ``achieving a goal''.
Even with the task of mathematical reasoning, where a rule-based reward model is used to verify whether the answer is correct or not, we do not have a mathematical oracle that tells us the outcome is correct or not, but we have to generate the reward in a data-driven manner.

Finally, the discount factor $\gamma$ determines the preference over response conciseness. When setting it to $1$, it means generating a correct response containing $300$k tokens would be equally good to another correct response using $300$ tokens (e.g., in the thinking mode when we prioritize the correctness of the final answer of a challenging math question). When setting it to a number smaller than $1$, it means we would prefer more concise or shorter responses to finish a given task.

\begin{table}[h]
% \fontsize{8}{7}\selectfont
\centering
\caption{\small LLM generation as an MDP\textbackslash R}
\begin{tabular}{ll}
\toprule
\textbf{Component} & \textbf{Interpretation} \\
\midrule
$\mathcal{S}$ (State) & Current sentence \\
$\mathcal{A}$ (Action) & Tokens (or their combinations) \\
$\mathcal{P}$ (Transition) & Concatenation of tokens \\
$\rho_0$ (Initial state distribution) & Prompt / Query distribution \\
\textcolor{brown}{$\mathcal{R}$ (Reward)} & \textcolor{brown}{\textit{Data-Driven}}\\
$\gamma$ (Discount factor) & $\leq 1$ (e.g., no discount vs. brevity preference) \\
\bottomrule
\end{tabular}
\label{tab:2}
\end{table}

Table~\ref{tab:2} summarizes the MDP components of LLM generation, with a highlighted (missing) reward function that has to be generated in a data-driven approach. In the next section, we will revisit the classical methods in the RL literature and draw inspiration from the classics in solving MDP\textbackslash R problems.

\subsection{MDP\textbackslash R: Markov Decision Processes without Reward Function}
In MDPs, the learning objective is to maximize cumulative reward over decision steps. However, in an MDP\textbackslash R, how to effectively optimize the policies without a reward function?
In RL literature, we can learn from a \textbf{behavior dataset} in those MDP\textbackslash R settings. 
\paragraph{Motivations and Practices of MDP\textbackslash R: Learning from Behavior Datasets}
In many real-world tasks, reward signals are difficult to specify. For example, in early autonomous driving systems such as ALVINN~\citep{pomerleau1988alvinn}, the learning objective is to mimic human driving behavior—a goal that is inherently hard to formalize as a reward function. More generally, in imitation learning setups~\citep{hayes1994robot}, behavior datasets serve as a direct and expressive means of specifying desired behaviors, without the need for manually crafted reward functions. This difficulty in defining explicit reward signals is also evident in complex robotic skill learning~\citep{peng2018deepmimic}, where behaviors such as agile locomotion or acrobatic motions are more easily demonstrated than described through rewards. In the context of LLM alignment~\citep{bai2022constitutional}, properties such as helpfulness, harmlessness, and summarization quality are similarly challenging to quantify through reward functions alone~\citep{stiennon2020learning, ouyang2022training, bai2022training}.

Beyond ill-defined reward settings, behavior datasets are also valuable in problems where reward functions are well-defined but sparse. In such scenarios, learning from behavior can substantially aid exploration. A canonical example is the game of Go, where the objective of "winning" is clearly defined, yet extremely difficult to achieve from random play due to the sparsity of the reward. Systems such as AlphaGo, AlphaStar, and OpenAI Five successfully leveraged expert demonstrations or replay data from human players to initialize and guide policy learning~\citep{silver2016mastering,vinyals2019grandmaster,berner2019dota}. Similarly, in robotics control tasks with sparse but well-defined success-based reward signals, incorporating expert demonstrations can significantly enhance sample efficiency and guide exploration. This class of techniques is broadly known as Learning from Demonstrations (LfD)~\citep{nair2018overcoming,hester2018deep}.

\paragraph{Methods for MDP\textbackslash R: Imitation Learning and Inverse Reinforcement Learning}
In general, approaches to learning from behavior can be broadly categorized into two classes: Imitation Learning (IL) and Inverse Reinforcement Learning (IRL). Both can be interpreted as instances of behavioral distribution matching, where the goal is to align the learned policy’s behavior distribution with that of the expert~\citep{ghasemipour2020divergence, ke2021imitation}. A common assumption underlying both IL and IRL is the availability of the environment dynamics, allowing for potentially unlimited interactions with the environment through rollouts.

In contrast, their offline counterparts --- namely Offline IL and Offline IRL --- operate under the constraint that the environment dynamics are unknown and no additional interaction is possible~\citep{jiang2020offline,jarrett2020strictly,yu2023offline}. This offline setting introduces significant challenges, most notably the inability to explore counterfactual behaviors that are not present in the demonstration dataset~\citep{fujimoto2019off}. As a result, the learned policy is limited to the support of the existing data, which can hinder its ability to improve beyond what is demonstrated~\citep{zolna2020offline}.

In light of the challenges discussed above, particularly the issues of distributional shift and compounding errors in the offline setting, we now turn to practical algorithms for IL and IRL that explicitly leverage access to the environment dynamics. These methods exploit interactions with the environment to mitigate error accumulation and achieve more robust policy learning.
\begin{table}[t!]
\fontsize{8}{5}\selectfont
\centering
\caption{\small Summarizing difference in problem settings of RL, Offline-RL, Imitation Learning (IL), Inverse-RL, Offline Inverse-RL (Offline IRL), Learning from Demonstrations (LfD), and Preference-based RL.}
\begin{tabular}{l|c|c|c|c|c}
\toprule
\textbf{Problem} & \textbf{External} & \textbf{External} & \textbf{Learned} & \textbf{Behavior} & \textbf{Examples} \\
\textbf{Settings} & \textbf{Dynamics} & \textbf{Reward} & \textbf{Reward} & \textbf{Dataset} & \textbf{Solvers} \\
\textbf{} & \textbf{Model} & \textbf{Model} & \textbf{Model} & \textbf{} & \\
\midrule
RL & $\cmark$ & $\cmark$ & $\xmark$ & $\xmark$ & {PPO~\citep{schulman2017proximal},} \\
 &  &  &  &  & {TD3~\citep{fujimoto2018addressing},} \\
 &  &  &  &  & {SAC~\citep{haarnoja2018soft}} \\
\midrule
Offline-RL & $\xmark$ & $\xmark$ & $\cmark$ or $\xmark$ & $\cmark$ & {BC~\citep{pomerleau1991efficient},} \\
 &  &  &  &  & {CQL~\citep{kumar2020conservative},} \\
 &  &  &  &  & {WGCSL~\citep{yang2022rethinking}} \\
\midrule
Imitation & $\cmark$ & $\xmark$ & $\xmark$ & $\cmark$ & {BC~\citep{pomerleau1991efficient},} \\
 &  &  &  &  & {AOC~\citep{sun2023accountable},} \\
 &  &  &  &  & {GAIL~\citep{ho2016generative}} \\
\midrule
Inverse-RL & $\cmark$ & $\xmark$ & $\cmark$ & $\cmark$ & {BC~\citep{pomerleau1991efficient},} \\
 &  &  &  &  & {AIRL~\citep{fu2017learning}} \\
\midrule
Offline-IRL & $\xmark$ & $\xmark$ & $\cmark$ & $\cmark$ & {BC~\citep{pomerleau1991efficient},} \\
 &  &  &  &  & {AOC~\citep{sun2023accountable},} \\
 &  &  &  &  & {SBIL~\citep{jarrett2020strictly}} \\
\midrule
LfD & $\cmark$ & $\cmark$ & $\xmark$ & $\cmark$ & {DQNfD~\citep{hester2018deep},} \\
 &  &  &  &  & {DDPGfD~\citep{nair2018overcoming},} \\
 &  &  &  &  & {AlphaStar~\citep{vinyals2019grandmaster}} \\
\midrule
Preference- & $\cmark$ & $\xmark$ & $\cmark$ & Preference & {CPL~\citep{hejna2023contrastive},} \\
based RL &  &  &  &  & {T-REX~\citep{brown2019extrapolating},} \\
 &  &  &  &  & {RLHF~\citep{christiano2017deep,ouyang2022training},} \\
 &  &  &  &  & {DPO~\citep{rafailov2023direct}} \\
\bottomrule
\end{tabular}
\label{tab:alihan-dan}
\end{table}
\subsection{Practical IL and IRL Algorithms}
To make these ideas concrete, we now review practical algorithmic implementations of IL and IRL, focusing on how access to environment dynamics helps address the limitations of purely offline learning. We begin with the most basic form of imitation learning, Behavior Cloning (BC), which requires only demonstration data and no environment interaction, and then discuss more advanced approaches that incorporate rollouts and interactions for distribution matching.

In IL, the objective is to recover the behavior of an expert policy $\pi_\beta$ using a parameterized learner policy $\pi$. The most straightforward approach of IL is BC~\citep{pomerleau1988alvinn}, which instantiates the imitation through supervised learning.
\begin{graybox}
\small
\paragraph{Behavior Cloning~\citep{pomerleau1988alvinn}} 
A demonstrative decision dataset is collected from a behavior policy $\pi_\beta$. Denoting the state-action pairs in the dataset as $(s_i, a^*_i)\sim\mathcal{D}$, the BC method learns a policy through a supervised learning objective:
\begin{equation*}
\small
    \pi_\mathrm{BC} = \arg\max_\pi \mathbb{E}_{(s_i,a_i)\sim\mathcal{D}_\mathrm{demo}} \log(\pi(a_i|s_i))
\end{equation*}
% that generates the dataset.
\end{graybox}
Despite its simplicity and minimal requirements on the behavioral data (i.e., only needing state and action pairs but nothing else), its offline nature leads to a fundamental challenge --- known as the \textit{distributional shift}: in evaluation, the state distribution is sampled from rolling out the learned policy $\pi$, rather than the behavior policy $\pi_\beta$ that generates the dataset.
The expected number of mistakes made by the learned policy $\pi$ based on such an expert decision dataset can be denoted as
\begin{equation}
\ell(\pi) = \mathbb{E}_{p_\pi(\tau)} \left[ \sum_{t=0}^T \mathbbm{1}(\pi(s_t)\ne a^*_t) \right]    
\end{equation}
Then we have the following theorems:
\begin{theorem}[Behavior Clone Error Bound. \citep{ross2011reduction}]
\label{theorem:1}
     If $\pi$ is trained via empirical risk minimization on $s_t\sim p_{\pi_\beta}(\tau)$ and optimal labels $a_t^*$, and attains generalization error $\epsilon$ on $s_t \sim p_{\pi_\beta}(\tau)$, then $\ell(\pi)\le C+T^2 \epsilon$ is the best possible bound on the expected error of the learned policy.
\end{theorem}
\begin{remark}[Compounding Error.]
    An intuitive interpretation of this quadratic relationship between the error bound and the generalization error is that those errors aggregate along the trajectory. i.e., whenever the learned policy makes a mistake, it tends to make more mistakes from then on as that action is not optimal and will lead to other out-of-distribution states, which will lead to further mistakes.
\end{remark}

In order to alleviate the challenge of compounding error we discussed above, IL considers the setting where a dynamic model is available during learning. 
The objective of IL is to learn from a (decision) demonstration dataset, with access to a dynamics model --- such that the \textbf{current policy can be rolled out in the real environment}. Intuitively, with such a dynamics model, the optimization objective will no longer be $s_t\sim p_{\pi_\beta}(\tau)$ but could be $s_t\sim p_{\pi}(\tau)$ --- \textbf{the distributional shift problem can be alleviated.} It has been shown in the literature that having access to a \textit{dynamics model} is essential in controlling the error bound.~\citep{ross2011reduction}

\begin{theorem}[DAgger Error Bound, \citep{ross2011reduction}]
\label{theorem:2}
    If $\pi$ is trained via empirical risk minimization on $s_t\sim p_{\pi}(\tau)$ and optimal labels $a_t^*$, and attains generalization error $\epsilon$ on $s_t\sim p_{\pi}(\tau)$, then $\ell(\pi)\le C+T \epsilon$ is the best possible bound on the expected error of the learned policy.
\end{theorem}
\begin{remark}
    This requires the additional assumption of being able to access the behavior (expert) policy $\pi_\beta$ actively to acquire the expert for those roll-out trajectories generated by $\pi$ . 
\end{remark}

Beyond supervised imitation, adversarial imitation learning (AIL) methods such as GAIL~\citep{ho2016generative} formulate imitation as a distribution matching problem using adversarial training, drawing direct inspiration from generative adversarial networks (GANs). These methods introduce a discriminator to distinguish expert from learner behavior, and optimize the policy to fool this discriminator.

While specific AIL methods such as GAIL provide effective practical algorithms, a more general understanding can be achieved by viewing them through the lens of $f$-divergence minimization~\citep{nowozin2016f}. This perspective reveals that many AIL variants are in fact instantiations of a unified framework, differing only in the choice of divergence function. The following formulation, proposed by~\citet{ghasemipour2020divergence}, provides a general min-max optimization structure underlying these methods:

\begin{graybox}
\small
\paragraph{$f$-divergence Adversarial Imitation Learning (\citet{ghasemipour2020divergence}} 
The general adversarial imitation learning problem can be formalized as the following min-max objective:
\begin{equation}
\label{eqn:fgan}
    \min_{\pi} \max_{T_\omega} \mathbb{E}_{(s,a)\sim\mathcal{D}_\mathrm{demo}}[T_\omega(s,a)] - \mathbb{E}_{(s,a)\sim\pi}[f^*(T_\omega(s,a))]
\end{equation}
 where $f:\mathbb{R}^+\mapsto\mathbb{R}$ is a convex, lower-semicontinuous
function, and it defines a statistical divergence between distribution $P,Q$ with density function $p,q$ as: $D_f(P||Q) = \large\int_x q(x) f\left(\frac{p(x)}{q(x)}\right)dx$, and $f^*$ is the conjugate of $f$, defined as $f^* = \sup_{u\in \mathrm{dom}_f}\{ut - f(u)\}$. Practically, Equation (\ref{eqn:fgan}) can be solved through iterative optimizing
\begin{equation}
% \label{eqn:fgan-inner}
    \max_{T_\omega} \mathbb{E}_{(s,a)\sim\mathcal{D}_\mathrm{demo}}[T_\omega(s,a)] - \mathbb{E}_{(s,a)\sim\pi}[f^*(T_\omega(s,a))]
\end{equation}
and
\begin{equation}
% \label{eqn:fgan-inner}
    \max_{\pi} \mathbb{E}_{\tau\sim\pi}[\sum_{t}f^*(T_\omega(s_t,a_t))]
\end{equation}
\end{graybox}
Using $\rho$ to denote the state-action visitation frequency\footnote{\citet{ni2021f} discussed when only state visitation frequency is available.}, Table~\ref {tab:AIL_methods} elaborates on how different choices of $f$ lead to different practical implementations of the AIL approach.
\begin{table}[h!]
% \fontsize{8}{7}\selectfont
\centering
\caption{\small Different $f$-divergences used in Adversarial IRL methods}
\begin{tabular}{llll}
\toprule
\textbf{Method} & \boldmath$f(u)$ & \textbf{Divergence} & \boldmath$D_f(\rho^{\mathrm{demo}} || \rho^{\pi})$ \\
\midrule
AIRL~\citep{fu2017learning} & $- \log u$ & Reverse KL & $\mathrm{KL}(\rho^\pi || \rho^{\mathrm{demo}})$ \\
GAIL~\citep{ho2016generative} & $-(u + 1) \log \frac{1+u}{2} + u \log u$ & Jensen-Shannon & $\mathrm{JS}(\rho^\pi || \rho^{\mathrm{demo}})$ \\
FAIRL~\citep{ghasemipour2020divergence} & $u \log u$ & Forward KL & $\mathrm{KL}(\rho^{\mathrm{demo}} || \rho^\pi)$ \\
% $\alpha$-IRL~\citep{} & $\frac{u^{1-\alpha} - (1-\alpha)u - a}{\alpha(\alpha-1)}$ & $\alpha$-divergence & $D_\alpha(\rho^{\mathrm{demo}} || \rho^\pi)$ \\
\bottomrule
\end{tabular}
\label{tab:AIL_methods}
\end{table}
% \begin{itemize}
%     \item AIRL: $f(u) = - \log (u)$ ; ~~\qquad\qquad \qquad\qquad$D_f(\rho^\mathrm{demo} || \rho^\pi) = \mathrm{KL}(\rho^\pi || \rho^\mathrm{demo}) $
%     \item GAIL: $f(u) = -(u + 1) \log \frac{1+u}{2} +  u \log u$; \quad$D_f(\rho^\mathrm{demo} || \rho^\pi) = \mathrm{JS}(\rho^\pi|| \rho^\mathrm{demo} ) $
%     \item FAIRL: $f(u) = u\log (u)$; ~~\qquad\qquad\qquad\qquad$D_f(\rho^\mathrm{demo} || \rho^\pi) = \mathrm{KL}(\rho^\mathrm{demo} || \rho^\pi)$
%     \item $\alpha$-IRL: $f(u) = \frac{u^{1-\alpha} - (1-\alpha)u -a}{\alpha(\alpha-1)}$; ~~\quad\qquad\qquad $D_f(\rho^\mathrm{demo} || \rho^\pi) = D_\alpha(\rho^\mathrm{demo} || \rho^\pi)$
% \end{itemize}

Importantly, the difference between those algorithms also highlights the difference between IL and IRL: in IL, the learning objective is to directly recover the expert behavior by imitating it, whereas in IRL, a reward model is learned from the behavior datset, such that maximizing accumulated return predicted by such a learned reward will induce the behavior policy. 

\begin{mdframed}[innertopmargin=0pt,leftmargin=0pt, rightmargin=0pt, innerleftmargin=10pt, innerrightmargin=10pt, skipbelow=0pt]
\textbf{\textcolor{brown}{Take-aways} (1). The access to environmental dynamics is essential.} It enables distributional matching besides the offline BC objective, hence alleviating the distributional shift and compounding error problems. \textbf{(2). Reward Models in IRL are not unique.} Different assumptions lead to different reward models. We will demonstrate such a point in Section~\ref{sec:rm_in_the_wild}.
\end{mdframed}

\section{Optimizing LLMs beyond Imitation: Why do we Need Neural Reward Models}
\subsection{LLMs as Language Imitators}
Given a training corpus, the LLM pre-training and Supervised Fine-Tuning are both performing next token prediction tasks~\citep{radford2018improving}. Empirically we know that when data, compute, and model scales, those pre-trained models begin to obtain emergent abilities of understanding and comprehending zero-shot complex tasks~\citep{kaplan2020scaling,wei2022emergent,kojima2022large}. From the perspective of RL from behavior dataset, such pre-training and SFT processes are imitating the behavioral datasets through BC~\citep{srivastava2022beyond,sun2024inverse}.
\begin{figure}[h!]
    \centering
    \includegraphics[width=1.0\linewidth]{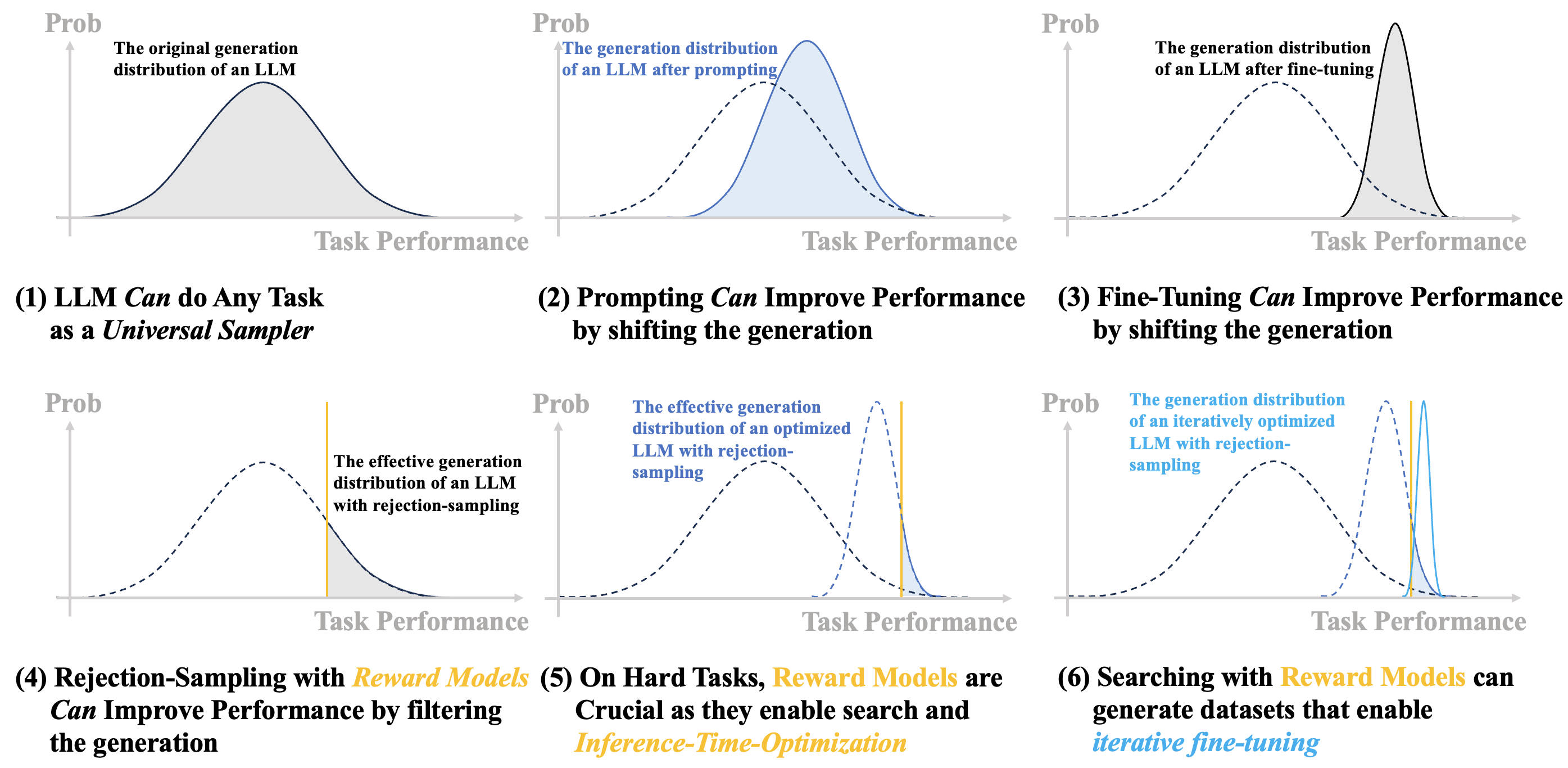}
    \vspace{-0.2cm}
    \caption{\small \textit{A comparison of different LLM generation optimization approaches.} The first row represents (1) direct generation, (2) prompt optimization, (3) Supervised Fine-Tuning (SFT) on a high-quality dataset. The second row represents methods that leverage reward models (i.e., the IRL approach): (4) reward models can be used to filter out low-quality generations, (5-6) reward models can be combined with prompt optimization or fine-tuning methods to improve the generation quality. \textbf{Only reward models enable inference time optimization.}}
\label{fig:teaser_llm_imitator_irl}
\end{figure}
In panel (1) of Figure~\ref{fig:teaser_llm_imitator_irl}, we illustrate the LLM generation as a distribution on different task-specific performance. The objective of LLM optimization is to shift the performance distribution to have higher scores on average (i.e., shift the distribution to the right side on the x axis).
Since LLMs are conditional generators, asking the same question with different prompting strategies can lead to a huge difference in task performance. For instance, the chain of thought prompting was known to be very successful in improving the model's ability to solve reasoning tasks~\citep{wei2022chain} --- as illustrated by panel (2) of Figure~\ref{fig:teaser_llm_imitator_irl}. While simple and effective, engineering prompts can be costly since those optimal prompt strategies are always model- and query-dependent~\citep {sun2023query,yang2023large}.
An alternative approach to prompting --- when having a high-quality demonstration dataset --- is to conduct SFT on those relatively small datasets. LLMs can do decently well on few-shot learning~\citep{brown2020language}.

Given that behavior cloning, fine-tuning, and prompting have shown strong empirical performance in aligning LLMs, one may ask: why do we need Inverse RL for alignment? Why do we need explicit reward models? In the following, we will discuss 3 motivations for learning explicit reward models (i.e., optimizing LLMs using IRL).

\subsection{Reward Models in Conversational AI: Toward Practical and Scalable Learning from Human Feedback}
In real-world deployments of chat models, it is common to collect preference-based feedback from users to improve response quality~\citep{ouyang2022training}. While asking users to directly provide demonstrations (i.e., ideal responses) would be desirable in principle, doing so at scale is prohibitively expensive and cognitively demanding.

Practically, users and annotators are often more capable at discriminative tasks, such as choosing the better of two options, than at generative tasks such as writing full responses. This aligns with findings in prior work~\citep{brown2019extrapolating}, which show that collecting preference data is a more practical and scalable way to build reward models and achieve super-demonstrator performance. 

As a result, learning reward models from preference data has become a core technique in large-scale RLHF pipelines, enabling practical supervision without the need for expert-level demonstrations.

\begin{mdframed}[innertopmargin=0pt,leftmargin=0pt, rightmargin=0pt, innerleftmargin=10pt, innerrightmargin=10pt, skipbelow=0pt]
\textbf{\textcolor{brown}{Take-away}} Demonstration data is not always available. Preference data is more scalable and practical. IRL enables flexible acquisition of data and annotation for learning from behavioral datasets and implicit metrics.
\end{mdframed}

\subsection{Reward Models in Mathematical Reasoning: Learning Generalizable Reasoning Skills from Math}
The second case where we need a reward model is mathematical reasoning. 
In those tasks, conducting SFT on demonstrative datasets may improve accuracy on seen question types but often fails to generalize to consistent and effective reasoning patterns. This limitation stems from the inherent difficulty of capturing complex, compositional reasoning behaviors via static datasets and imitation.

In such tasks, reward models offer a flexible mechanism for generalization. Instead of directly mimicking demonstrations, LLMs can be optimized to \textbf{explore and discover high-reward reasoning trajectories}, guided by feedback encoded in the reward model. Systems such as DeepSeek-R1~\citep{guo2025deepseek} demonstrate that using rule-based or learned reward models enables LLMs to exhibit behaviors such as deep thinking, long chains of reasoning, and self-correction capabilities that are difficult to induce through imitation alone.

\begin{mdframed}[innertopmargin=0pt,leftmargin=0pt, rightmargin=0pt, innerleftmargin=10pt, innerrightmargin=10pt, skipbelow=0pt]
\textbf{\textcolor{brown}{Take-away}} Reward models enable generalization in math reasoning tasks. Data-driven reward functions allow RL algorithms to discover and reinforce generalizable reasoning behaviors such as deep thinking, long chain-of-thought, and self-correction.
\end{mdframed}

\subsection{Reward Models for Test-Time Optimization}
A unique advantage of reward models lies in their ability to support \textbf{inference-time (test-time) optimization}. While prompting and SFT can enhance task-specific performance, these approaches typically operate offline, and the improvements achieved during training cannot be adapted during test-time generation.

In classical reinforcement learning tasks, not every setting requires inference-time optimization. In relatively simple environments such as MuJoCo locomotion or Atari games, inference often consists of a single forward pass through a trained policy network. In contrast, more complex tasks such as Go require test-time optimization, where search-based planning guided by value estimators is critical for achieving superhuman performance.

Similarly, in LLM generation tasks, reward models can be used to enable inference-time optimization. For example, given a trained reward model (illustrated as the golden vertical line in panel (4) of Figure~\ref{fig:teaser_llm_imitator_irl}), candidate generations can be evaluated and low-quality outputs filtered out during inference. In high-stakes domains such as mathematical reasoning or instruction-following, this enables hybrid strategies that combine prompting, supervised fine-tuning, and reward-model-based optimization to improve test-time performance.

\begin{mdframed}[innertopmargin=0pt,leftmargin=0pt, rightmargin=0pt, innerleftmargin=10pt, innerrightmargin=10pt, skipbelow=0pt]
\textbf{\textcolor{brown}{Take-away}} Reward models enable inference-time (test-time) optimization by scoring and filtering generated outputs. This allows LLMs to adaptively select high-quality responses during deployment, analogous to test-time planning in classical RL tasks.
\end{mdframed}

\section{From Real World Evidences to Alignment: Practical IRL via Reward Modeling}
\label{sec:rm_in_the_wild}

Alignment fundamentally concerns ensuring that machine behavior is consistent with the real world and its implicit objectives. The world is rich with observable signals: demonstrations, preferences, behaviors, and choices, which reflect underlying goals and constraints. These signals serve as the evidence upon which alignment should be based.

Achieving such alignment requires learning from real-world data rather than relying solely on manually specified objectives. In this context, a practical and principled approach is to infer reward functions from observed behavior. This allows us to translate real-world evidence into actionable objectives for learning and decision-making.

Crucially, real-world data is often noisy, partial, or biased. Human behavior may be suboptimal, inconsistent, or poorly articulated. Nevertheless, such data remains one of the most informative sources for guiding LLM post-training. By extracting structure and intent from this evidence, we can better align models with actual human goals, even when those goals are not explicitly stated.

This section focuses on how post-training can be operationalized through IRL, particularly by leveraging observed behavior to build effective reward models (RMs).

\subsection{Reward Modeling from Preference Feedback}
\label{sec:pref_rms}
\paragraph{From PPO to DPO: Reinforcement Learning from Human Feedback (RLHF) as IRL}
Reinforcement Learning from Human Feedback (RLHF) has become a standard paradigm for aligning large language models~\citep{stiennon2020learning,ouyang2022training,bai2022training,bai2022constitutional}. The core idea involves learning a reward model from human preference data and then using this model to guide policy optimization.

The training data typically consists of pairwise preferences over model outputs: $\mathcal{D}_\mathrm{pref} = \{(x_i, y_i^+, y_i^-)\}_{i=1}^N$, where $x_i$ is a query and $y_i^+, y_i^-$ denote the preferred and less-preferred responses, respectively. To convert these comparisons into scalar reward signals, models such as Bradley-Terry~\citep{bradley1952rank} or logistic preference models are employed. These assign relative scores to responses such that $r(y_i^+) > r(y_i^-)$, enabling reward modeling through pairwise loss functions.
\begin{graybox}
\small
\paragraph{RLHF with Bradley-Terry Reward Models~\citep{christiano2017deep}} 
In standard RLHF, a reward model $r_\theta: (x, y) \mapsto \mathbb{R}$ is trained to reflect human preferences. Given a dataset of pairwise comparisons $\mathcal{D}_\mathrm{pref} = \{(x_i, y_i^+, y_i^-)\}$, where $y_i^+$ is preferred over $y_i^-$ for query $x_i$, the reward model is optimized via the Bradley-Terry likelihood:
\[
\mathcal{L}_{\mathrm{BT}}(\theta) = \sum_{i=1}^N \log \sigma\left(r_\theta(x_i, y_i^+) - r_\theta(x_i, y_i^-)\right),
\]
where $\sigma(z) = \frac{1}{1 + e^{-z}}$ is the logistic function. This encourages the model to assign higher reward values to preferred responses.
\end{graybox}
% The learned reward model is then frozen, and used to supervise policy optimization via Proximal Policy Optimization (PPO)~\citep{schulman2017proximal}, which maximizes expected predicted reward while constraining the updated policy to stay close to a reference policy under a KL-divergence constraint. We will elaborate on the policy optimization steps in Section~\ref{sec:policy_optimization}.

% While PPO performs online policy updates with sampled trajectories and advantage-weighted gradients, DPO directly optimizes the policy to satisfy pairwise preference constraints under a KL-regularized objective.
The learned reward model is then frozen and used to supervise policy optimization via Proximal Policy Optimization (PPO)~\citep{schulman2017proximal}. PPO maximizes the expected reward predicted by the learned reward model while constraining the updated policy to remain close to a reference policy using a KL-divergence penalty. We elaborate on policy optimization details in Section~\ref{sec:policy_optimization}.

In contrast, Direct Preference Optimization (DPO)~\citep{rafailov2024direct} sidesteps the explicit reward modeling and trajectory sampling steps altogether. Instead, it directly optimizes the policy to satisfy pairwise preference constraints derived from human feedback, using a KL-regularized classification-style objective over prompt-response pairs. This leads to a simpler and empirically more stable training process compared to PPO-based RLHF pipelines.
\begin{graybox}
\small
\paragraph{Direct Preference Optimization (DPO)~\citep{rafailov2024direct}} 
DPO reinterprets preference-based RLHF as a probabilistic inference problem. The key idea is to start from a latent reward function and derive a policy learning objective that avoids explicitly modeling the reward.

Assume a latent reward function $r(x, y)$ governs human preferences via a Bradley-Terry model:
\[
P(y^+ \succ y^- \mid x) = \frac{\exp(r(x, y^+))}{\exp(r(x, y^+)) + \exp(r(x, y^-))}.
\]
The optimal policy $\pi^*$ can be derived~\footnote{For derivation details, please refer to \citep{peters2007reinforcement,wang2018exponentially,yang2022rethinking,peng2019advantage}.} as:
\[
\pi^*(y \mid x) = \frac{\exp(\beta r(x, y))}{Z(x)}, \quad \text{where } Z(x) = \sum_{y'} \exp(\beta r(x, y')).
\]
Then the reward difference can be rewritten in terms of the optimal policy:
\[
r(x, y^+) - r(x, y^-) = \frac{1}{\beta} \left[ \log \pi^*(y^+ \mid x) - \log \pi^*(y^- \mid x) \right].
\]
DPO approximates $\pi^*$ with a learnable policy $\pi_\phi$, and directly maximizes the likelihood of human preferences:
\[
\mathcal{L}_{\mathrm{DPO}}(\phi) = \sum_{i=1}^N \log \sigma\left( \beta \left[ \log \pi_\phi(y_i^+ \mid x_i) - \log \pi_\phi(y_i^- \mid x_i) \right] \right),
\]
where $\sigma(z) = \frac{1}{1 + e^{-z}}$ is the logistic function.
\end{graybox}

This objective avoids reward model training by directly adjusting the policy's log-probabilities to match observed preferences, while implicitly capturing the reward structure through relative likelihoods.

% From the perspective of IRL, both approaches are IRL methods, as reward modeling is either explicitly or implicitly engaged in the learning process. The conventional 2-stage RLHF methods explicitly learn the reward models. On the other hand, DPO circumvents the reward model parameterization and uses policy models to implicitly express the reward model instead.

% From the perspective of IRL, both RLHF and DPO can be viewed as IRL methods, as they involve inferring preferences or underlying objectives from human feedback. 

% This distinction invites further comparison between the two paradigms. While reward-model-based RLHF offers flexibility in decoupling supervision signals from policy training, it often suffers from instability and reward misspecification. DPO, by integrating preference supervision directly into the policy training objective, avoids reward modeling computational burden but introduces constraints on expressivity and interoperability. In recent advances, \citet{xu2024dpo,ivison2024unpacking} both demonstrate the general superiority of the explicit reward modeling approach with PPO over DPO, as long as the hyperparameters of PPO are appropriately adjusted. However, stabilizing the training of PPO is always non-trivial in practice~\citep{rafailov2024direct}, and DPO can be more robust to overoptimization as compared to PPO~\citep{ivison2024unpacking}. Recent works tries to combine both approaches to integrate the pros form both sides~\citep{zhong2024dpo}. 

From the perspective of IRL, both RLHF and DPO can be viewed as IRL methods, as they involve inferring preferences or underlying objectives from human feedback. Recent studies~\citep{xu2024dpo,ivison2024unpacking} have shown that reward-model-based RLHF can outperform DPO when PPO hyperparameters are properly tuned. However, stabilizing PPO remains non-trivial in practice~\citep{rafailov2024direct}, and DPO tends to be more robust to overoptimization~\citep{ivison2024unpacking}. Recent work has explored hybrid approaches that aim to combine the strengths of both methods~\citep{zhong2024dpo}. Alternatively, one may consider iterative DPO as an online variant of the original DPO, and achieve improved learning efficiency while sustaining high stability~\citep{xiong2023gibbs}. Recent advances given in \citet{shi2025understanding} further provide theoretical insights on the priority of different approaches.
% \begin{mdframed}[innertopmargin=0pt,leftmargin=0pt, rightmargin=0pt, innerleftmargin=10pt, innerrightmargin=10pt, skipbelow=0pt]
% \textbf{\textcolor{brown}{Take-away}} DPO is superior in stabilizing the alignment process without need of hyperparameter tuning. When properly tuned, PPO can outperform DPO on general tasks. The choice of algorithm for a real-world task should be depend on the computational resources and performance-sensitivity.
% \end{mdframed}
\begin{mdframed}[innertopmargin=0pt,leftmargin=0pt, rightmargin=0pt, innerleftmargin=10pt, innerrightmargin=10pt, skipbelow=0pt]
\textbf{\textcolor{brown}{Take-away}} DPO offers superior training stability and requires less hyperparameter tuning, making it a robust choice for alignment. In contrast, PPO with explicit reward modeling can outperform DPO when carefully tuned. The selection between the two should be guided by task sensitivity and available computational resources.
\end{mdframed}

% In the seminal works of RLHF for LLM alignment~\citep{ouyang2022training,bai2022constitutional,bai2022training,stiennon2020learning}, datasets in the form of preference annotations are needed. $\mathcal{D}_\mathrm{pref} = \{x_i, y_i^+, y_i^-\}_{i\in[N]}$, where $y_i^+$ and $y_i^-$ are the preferred and dis-preferred responses given input $x_i$. Models such as Bradley-Terry~\cite{bradley1952rank} are then used to convert ranking feedback into absolute scores to serve as reward signals.

% \paragraph{Rethinking the Foundation of Preference-based Reward Modeling}
% In the seminal papers of RLHF~\citep{christiano2017deep}, we know that we should use the Bradley-Terry (BT) models~\citep{bradley1952rank} to transfer pairwise preference annotations into scalar scores, such that the optimization objective of LLM alignment can be measured by the increase of such scalar scores~\citep{stiennon2020learning}. However, it was unclear why the BT model can be a solid choice, and there are alternative choices such as the Kahneman-Tversky model~\citep{ethayarajh2024kto} and general preference models~\citep{azar2024general}. 
In the seminal work on RLHF~\citep{christiano2017deep}, pairwise preference annotations are translated into scalar reward scores using the Bradley-Terry (BT) model~\citep{bradley1952rank}, enabling scalar supervision for aligning LLMs~\citep{stiennon2020learning}. However, the choice of BT has traditionally been made on heuristic grounds, and alternative preference models such as the Kahneman-Tversky ordinal model~\citep{ethayarajh2024kto} and general discrete choice models~\citep{azar2024general} have since been proposed.

% In \citet{sun2024rethinking}, a theoretical justification for the use of BT models in LLM alignment is first provided, and the paper highlights the important difference between BT parameter estimation~\citep{bradley1952rank} and BT-regression~\citep{springall1973response,bockenholt1988logistic}. The key insight is that BT reward models in RLHF rely on generalizable pre-trained LLM embeddings.
% Such a theoretical framework also explicitly justified the use of natural language embeddings for reward modeling~\citep{sun2025reusing,shen2025reviving,sun2023query}. 

% Furthermore, \citet{sun2024rethinking} introduced the objective of \textit{order consistency} in preference-based reward modeling --- the order consistency means the absolute value of the learned reward model is not as important as their relative orders, and this is because in downstream tasks such as Best-of-N optimization, only the order of responses matter. While BT model can achieve order consistency, it has been shown to be not necessary. The classification models --- using preferred responses as positive samples, and dispreferred responses as negative samples --- can also achieve the order consistency objective.
\citet{sun2024rethinking} provides a formal justification for the use of BT models in LLM alignment. Crucially, it distinguishes between \textit{BT parameter estimation}~\citep{bradley1952rank}, which assumes direct access to latent utilities, and \textit{BT regression}~\citep{springall1973response,bockenholt1988logistic}, which instead regresses reward scores from learned input representations. The key insight is that modern reward models operate in the \textit{embedding space} of pre-trained LLMs, making BT regression more appropriate. This perspective also provides a theoretical grounding for recent work that reuses or fine-tunes language model embeddings for reward modeling~\citep{sun2025reusing,shen2025reviving,sun2023query}.

Additionally, \citet{sun2024rethinking} introduces the notion of \textit{order consistency} as a more suitable learning objective in preference-based settings. That is, for tasks such as best-of-$N$ selection, the relative ordering of responses is more important than the absolute value of their scores. While the BT model satisfies this property, it is not uniquely suited for the task. Simpler alternatives, such as binary classification models that treat preferred responses as positives and dispreferred ones as negatives, can also achieve order-consistent objectives, and often perform better in the presence of noisy or ambiguous annotations.

\begin{mdframed}[innertopmargin=0pt,leftmargin=0pt, rightmargin=0pt, innerleftmargin=10pt, innerrightmargin=10pt, skipbelow=0pt]
\textbf{\textcolor{brown}{Take-away}} 
(1) The theoretical foundation of modern preference-based reward modeling is better captured by Bradley-Terry \textit{regression}~\citep{springall1973response} than classical BT estimation~\citep{bradley1952rank}.  
(2) Classification-based objectives can outperform BT models, particularly in the presence of annotation noise, while still preserving the crucial property of order consistency.
% (1). The mathematical foundation of preference-based reward modeling is the Bradley-Terry Regression model~\citep{springall1973response}, rather than the BT parameter estimation model~\citep{bradley1952rank}. (2). Using classification objective in preference-based reward modeling can achieve better performance than the BT reward models, especially when annotation quality is limited (i.e., high annotation noise cases.)
\end{mdframed}

\paragraph{Active Learning}

Building on this foundation, a central challenge in practical reward modeling is the efficient acquisition of preference annotations. Since reward models are trained to preserve the order of responses in embedding space, annotation strategies should prioritize comparisons that are most informative for determining rank. This has motivated the application of active learning in preference data collection.

Recent studies~\citep{muldrew2024active,mukherjee2024optimal} have proposed a variety of heuristic-inspired acquisition functions tailored to reward modeling. One commonly used approach is uncertainty sampling, which selects response pairs where the reward model is least confident in its preference. Another is maximum difference sampling, which selects pairs with the highest predicted reward gaps, under the assumption that such examples yield more reliable supervision. 

A more principled formulation of active preference learning is presented by \citet{shen2025reviving,feng2025pilaf}, which draws on tools from Fisher information theory and optimal experiment design. Instead of relying on heuristic acquisition functions, those papers propose to select query pairs that maximize the determinant of Fisher information with respect to the reward model parameters in the embedding space. 
\begin{graybox}
\small
\paragraph{Fisher-Information Guided Preference Annotation~\citep{feng2025pilaf,shen2025reviving}} 
Consider the linear BT regression models (on the embedding space), $r(x,y) = w^T\phi(x,y)$. The preference generation process of the $i$-th pair $h_i$ is
\[
h_i\sim \mathrm{Bernoulli}[\sigma[w^T(\phi(x,y_1)-\phi(x,y_2)]]
\]
Based on the theory fro generalized linear models, the maximum likelihood estimate $\hat{w}$ is asymptotically Gaussian distributed, with mean $w$ and covariance matrix $\mathcal{I^{-1}}$, where $\mathcal{I}$ denotes the Fisher Information (FI) matrix. 
\[
\mathcal{I} = \sum_{i=1}^I (\phi(x_i,y_{i,1})-\phi(x_i,y_{i,2}))^T(\phi(x_i,y_{i,1})-\phi(x_i,y_{i,2})) p_i(1-p_i)
\]
where $p_i = \sigma[w^T(\phi(x_i,y_{i,1})-\phi(x_i,y_{i,1}))]$. Following the classical methods \textit{Bayesian D-Optimality} design~\citep{chaloner1995bayesian}, preference annotations should prioritize those samples with the highest scores:
\[
\mathcal{S}_\mathrm{D-Opt} = |\mathcal{I}|
\]
\end{graybox}
This approach formalizes the goal of active learning as maximizing informativeness under a limited annotation budget. Importantly, their framework highlights an inherent \textit{exploration–exploitation trade-off}: selecting pairs that are highly uncertain (exploration) versus those that are expected to provide strong gradients for refining the current model (exploitation). Their method operates entirely in the embedding space, aligning well with recent theoretical insights about the structure of reward models and their reliance on pretrained LLM representations.

\begin{mdframed}[innertopmargin=0pt,leftmargin=0pt, rightmargin=0pt, innerleftmargin=10pt, innerrightmargin=10pt, skipbelow=0pt]
\textbf{\textcolor{brown}{Take-away}} 
\textbf{Building on top of linear BT models}, Fisher Information and optimal experimental design provide a theoretically grounded framework for active preference learning, highlighting the need to balance exploration and exploitation.
\end{mdframed}

\paragraph{Diverse Preferences and Personalization in Reward Modeling}

Beyond active sampling, another critical challenge in preference-based reward modeling is accounting for \textit{preference diversity}. In practice, human preferences vary significantly across users, tasks, and deployment settings. A single global reward model may fail to capture such heterogeneity, leading to poor generalization and potential misalignment with specific user intents~\citep{sorensen2024roadmap}.

To address this, recent works have explored personalized reward modeling through a variety of techniques. One direction is to explicitly learn user-specific latent variables that condition reward predictions~\citep{poddar2024personalizing,li2024personalized,kobalczyk2024few}. Others model reward distributions rather than point estimates, enabling uncertainty-aware reasoning over latent contextual factors~\citep{siththaranjan2023distributional}. Moreover, \citet{chakraborty2024maxmin} introduces a MaxMin training objective to align models with a diverse set of human preferences by optimizing worst-case reward performance across subgroups. Recent work of \citet{luo2025rethinking} innovates the usage of Principal Component Analysis (PCA) for lightweight personalized preference learning in the embedding space.
\begin{graybox}
\small
\paragraph{Decomposed Reward Models (DRMs)~\citep{luo2025rethinking}} 
Given a comparison triple $(x, y^+, y^-)$, where $y^+$ is the preferred response over $y^-$, the standard Bradley-Terry (BT) objective under vector representation space is:
\[
\max_{\mathbf{w}} \mathbb{E} \left[ \log \sigma\left( \mathbf{w}^\top \left( \phi(x, y^c) - \phi(x, y^r) \right) \right) \right],
\]
where $\phi(x, y)$ denotes the feature embedding of response $y$ conditioned on input $x$, $\mathbf{w}$ is a preference vector, and $\sigma(\cdot)$ is the sigmoid function.
Let $\Delta \phi_t = \phi(x_t, y^+_t) - \phi(x_t, y^-_t)$ and define the centered difference vector $z_t = \Delta \phi_t - \mathbb{E}[\Delta \phi]$. Consider the PCA over $z_t$, with the covariance matrix of feature differences:
\[
\Sigma = \mathbb{E}[z_t z_t^\top].
\]
By decomposing $\Sigma$, we obtain a set of orthogonal basis vectors in the embedding space that capture the main axes of variation in human preferences. Instead of modeling reward using a single vector $\mathbf{w}$, we define $d$ orthogonal reward heads:
\[
W = [\mathbf{w}_1, \dots, \mathbf{w}_d] \in \mathbb{R}^{h \times d},
\]
where each $\mathbf{w}_k$ corresponds to a principal direction extracted via PCA. The reward vector is then:
\[
\text{DRM}(x, y) = W^\top \phi(x, y) \in \mathbb{R}^d,
\]
which decomposes the original reward into $d$ interpretable components.
% When $\mathbf{w}^\top z_t$ is small (as in the initial phase of optimization), the BT objective is approximately:
% \[
% \max_{\mathbf{w}} \frac{1}{2} \mathbf{w}^\top \Sigma \mathbf{w}.
% \]
% This is equivalent to maximizing the Rayleigh quotient, whose solution is the principal eigenvector of $\Sigma$. Hence, \textbf{preference vectors that maximize the BT objective can be extracted via PCA}.
\end{graybox}
DRM is shown to be highly interpretable with preference attributes. Different reward heads specialize in different attributes, and the first head aligns with the majority preference, while others focus on different aspects. 
\begin{mdframed}[innertopmargin=0pt,leftmargin=0pt, rightmargin=0pt, innerleftmargin=10pt, innerrightmargin=10pt, skipbelow=0pt]
\textbf{\textcolor{brown}{Take-away}} 
\textbf{Building on top of linear BT models}, diverse human preferences can be expressed as vectors in the embedding space, and DRMs model them using a set of orthogonal basis vectors.
Such an approach offers a systematic way to understand human preferences by breaking complex preferences into interpretable parts.
\end{mdframed}

\subsection{Reward Modeling for Mathematical Reasoning}
\paragraph{Revisiting the History of LLM-based Math Reasoning Research}
LLMs have demonstrated strong competence in mathematical reasoning, yete methods for eliciting this capability have evolved rapidly. Early approaches centered around \textit{prompt optimization}, most notably Chain-of-Thought (CoT) prompting, which encourages step-by-step reasoning to improve final answer accuracy \citep{wei2022chain}. Subsequent variants such as zero-shot CoT \citep{kojima2022large}, self-consistency decoding \citep{wang2022self}, and Tree-of-Thoughts (ToT) prompting \citep{yao2023tree} expanded the space of inference-time strategies, showing that multi-step reasoning could be induced without altering model parameters. These techniques primarily operated at inference time and revealed that models possess latent reasoning abilities that can be surfaced with minimal intervention.

Despite their success, prompt-based methods are shown to be model-dependent~\citep{yang2023large}, and the black-box heuristics can not systematically detect or correct errors. This motivated a more structured paradigm based on search and planning with dense rewards~\citep{chan2024dense,lightman2023let}. Inspired by classical AI planning, these methods use algorithms such as Monte Carlo Tree Search (MCTS) to explore candidate reasoning paths, evaluating partial solutions with learned reward models or value functions~\citep{zhang2024rest}. By providing fine-grained feedback, these methods enable models to plan and search for optimized thoughts~\citep{pouplin2024retrieval}. While leveraging dense reward and MCTS may improve the math reasoning abilities~\citep{wang2023math}, it also introduces new challenges related to dense reward design, computational efficiency, and vulnerability to reward hacking~\citep{guo2025deepseek,gao2023scaling}.

More recently, the field has shifted toward reinforcement learning with verifiable rewards (RLVR), which leverages the fact that correctness in mathematical reasoning is often easily verifiable. Models such as DeepSeek-r1 have been trained using sparse but reliable correct-wrong signals, leading to significantly improved reasoning capabilities \citep{guo2025deepseek,jaech2024openai}. These models exhibit long, internally consistent chains of thought and frequently demonstrate behaviors such as self-reflection and backtracking. By directly optimizing for correctness, this paradigm departs from preference-based RLHF approaches and moves toward grounded, data-driven learning, without the need for neural reward models.

\paragraph{Evolving Understanding the Performance Gain from RL: Importance of Structure and Format}
Despite being framed as reinforcement learning, many recent advances in RLVR-based mathematical reasoning seem to benefit less from exploration in an RL environment, and more from the model’s alignment with effective response formats. Studies such as \citet{shao2025spurious,wang2025reinforcement} demonstrate that even spurious or minimal reward signals can significantly improve model performance. These improvements are often attributed not to the discovery of fundamentally new reasoning strategies but rather to the emergence of structured, verifiable, and execution-friendly templates, such as programmatic responses, long chain-of-thought derivations, or format-constrained contents.

In this light, RLVR can be viewed as a mechanism for internalizing template-level prompt optimization: unlike inference-time prompting strategies that guide structure externally, RLVR encourages the model to internalize such structures through training. This shift underscores a deeper insight --- that for complex reasoning tasks, such as math or code, the structure of the answer is as important as its content. The field may thus be entering a phase where format and reasoning inductive bias take center stage, even within the RL framework.

This convergence between RLVR and earlier prompt-based methods underscores the importance of revisiting prompt optimization, not merely as a heuristic, but as a principled framework for guiding model behavior through structured format design. In the following section, we examine recent advances on IRL-based prompt optimization for reasoning tasks, with a focus on the role of reward models in such a process.

% one-sample~\citep{wang2025reinforcement}, zero-sample~\citep{zhao2025absolute}, even randomized reward~\citep{shao2025spurious} can improve 

\paragraph{Revisiting Prompt Optimization: Building Proxy Verifiers from Prompting Experience}
% In prompt optimization research, literature has shown creativity in improving problem-solving capabilities of LLMs using diverse approaches, including CoT~\citep{kojima2022large}, ToT~\citep{yao2023tree}, in-context optimization~\citep{cui2024phaseevo}, multi-agent debate~\citep{smit2023should,du2023improving}, automated prompt optimization~\citep{pryzant2023automatic} such as LLM to optimize prompts~\citep{yang2023large,zhou2022large}.
% We refer \citet{li2025survey,cui2025automatic} as elaborated surveys for interested readers.

In the field of prompt optimization, recent work has explored a diverse range of strategies to enhance the problem-solving capabilities of large language models. These include CoT~\citep{kojima2022large}, ToT~\citep{yao2023tree}, in-context optimization~\citep{cui2024phaseevo}, multi-agent debate frameworks~\citep{smit2023should,du2023improving,liang2023encouraging}, task decomposition~\citep{khot2022decomposed,zhou2022least},and automated prompt search~\citep{pryzant2023automatic,guo2023connecting}, including approaches that leverage LLMs themselves to optimize prompts~\citep{zhou2022large,yang2023large}. 
For a comprehensive overview of this line of research, we refer readers to the recent surveys by \citet{li2025survey} and \citet{cui2025automatic}.

Among these methods, automated prompt optimization techniques that interact directly with the task environment by querying the LLM and receiving verifiable rewards often achieve strong performance without relying on explicit reward models. However, because they require repeated interactions with black-box models, these methods are computationally expensive and often impractical in real-world settings. Although effective, such approaches do not utilize reward models and do not exploit existing offline data to reduce interaction costs.

To address this limitation, Prompt-OIRL\citep{sun2023query} proposes a simple and cost-effective IRL-based method that reuses historical prompting trial-and-error experience to train a reward model for offline prompt evaluation and evaluation. Prompt-OIRL enables adaptive, query-dependent prompt selection without requiring additional calls to the LLM at inference time. This provides a practical and scalable solution to prompt optimization in settings where interaction cost is a bottleneck. The algorithm proceeds as follows:
\begin{graybox}
\small
\paragraph{Prompt Optimization with Offline IRL~\citep{sun2023query}} 
Prompt-OIRL builds upon prior work in prompt optimization by reusing experimental artifacts. Given an open-source query $q$ from a dataset $\mathcal{D}_{\mathrm{q}}$, a set of prompt candidates $p \in \mathcal{P}$ (either as prefix or suffix), and correctness labels $r^{(p,q)}$ indicating whether applying prompt $p$ to query $q$ yields a correct answer (i.e., $r^{(p,q)} = 1$ if correct, and $0$ otherwise), the method constructs a reward-labeled dataset for training.

In the \textit{Reward Modeling} phase, a reward model $\Upsilon^{(p,q)}_{\theta}$, parameterized by $\theta$, is trained to predict $r^{(p,q)}$ using a cross-entropy loss:
\begin{equation*}
\label{eqn:prompt-oirl}
    \mathcal{L}_{\text{CE}}(\theta;\mathcal{P},\mathcal{D}_\mathrm{q}) = -\mathbb{E}_{p\in\mathcal{P},q\in[\mathcal{D}_\mathrm{q}]} \left[
    r^{(p,q)} \log \sigma \left( \Upsilon_{\theta}{(p,q)} \right)
  + (1-r^{(p,q)}) \log \left(1-\sigma\left( \Upsilon_{\theta}{(p,q)}\right) \right)  \right]
\end{equation*}
Then in the \textit{Prompt Optimization} phase, the reward model $\Upsilon_{\theta}$ is used as a proxy to optimize prompt $p^*$ for any given query $q_i$:
\begin{equation*}
\label{eqn:prompt-oirl-opt}
    p^* = \arg\max_p \Upsilon_{\theta}{(p,q_i)}
\end{equation*}
\end{graybox}
Prompt-OIRL has shown significant improvement on mathematical reasoning tasks \textit{through reward modeling} for prompt optimization.
\begin{mdframed}[innertopmargin=0pt,leftmargin=0pt, rightmargin=0pt, innerleftmargin=10pt, innerrightmargin=10pt, skipbelow=0pt]
\textbf{\textcolor{brown}{Take-away}} 
Research in LLM-based mathematical reasoning has evolved from heuristic prompting to RLVR, yet recent findings suggest that RL's effectiveness often arises from structured response formats such as templating. Viewed as a general form of prompting, templating bridges RLVR and prompt optimization, highlighting the potential of automated methods like Prompt-OIRL for future progress.
% In the development of LLM-based mathematical reasoning, research has progressed from heuristic prompt optimization to more structured, data-driven approaches such as RLVR. However, recent findings suggest that the effectiveness of RL-based methods often stems from the emergence of structured response formats, such as templating and formatting. When viewed as a general form of prompting, templating reveals a conceptual connection between RLVR and prompt optimization. This observation highlights the potential of automated prompt optimization methods, such as Prompt-OIRL, to serve as a foundation for future advances in LLM-based reasoning research.
\end{mdframed}

\subsection{Reward Modeling from Demonstration Datasets}
While RLHF from preference learning and verifiable reward has demonstrated great success in aligning LLMs according to user intention or factual correctness, such data with binary identifiable labels is not universally applicable. Only a limited subset of tasks has a clear objective answer that is verifiable; for the majority, user-centered subjective evaluation is always essential.

Among subjective feedback types, preference data has become the most widely used, largely due to its scalability in practical annotation workflows. However, collecting high-quality preference annotations poses several challenges, including annotation noise and ambiguity~\citep{zheng2023secrets}, high labeling costs~\citep{guo2024direct,xiong2023gibbs,shen2025reviving}, and potential privacy concerns when sharing data with annotators~\citep{li2023privacy,pouplin2024retrieval}.

Beyond preferences, alternative feedback modalities such as demonstrations, scalar judgments, and critiques can often provide richer supervision, particularly in personalized or open-ended tasks~\citep{tandon2021learning,shi2022life,scheurer2022training,xiao2024leverage,li2024getting,chen2024self,sun2025openreview}. Recent work has explored Alignment from Demonstration (AfD)~\citep{sun2024inverse}, which learns reward models from expert demonstrations rather than pairwise comparisons. This direction aligns naturally with classical IRL literature. The most straightforward approach to AfD, like any IRL task, is BC. And recent works on AfD mainly work on going beyond such an approach.

Formally, \citet{sun2024inverse} revisited the occupancy matching problem of IRL~\cite{ho2016generative, ross2011reduction, fu2017learning, orsini2021matters} to enhance the performance of AfD. Using $\rho^\beta(s,a) = \pi_\beta(a|s)\sum_{t=0}\gamma^t \mathrm{Prob}(s_t = s|\pi_\beta)$ to denote the state-action occupancy measure of the behavior policy (i.e., the demonstrator), and $\rho^\pi(s,a)$ the state-action occupancy measure of the current policy. In the context of LLM generation, with $x$ the input query and $y = (y^{(0)},y^{(1)},...,y^{(T)}=\texttt{EOS} )$ the output response containing a maximum of $T+1$ tokens, the occupancy measure is
\begin{equation}
\begin{split}
    \rho^\pi(s_k,a_k) &= \rho^\pi(s_k = (x,y^{(0:k-1)}),a_k= y^{(k)})  \\
    &= \pi(a_k = y^{(k)}| s_k = (x,y^{(0:k-1)}))p(s_{k}) \\
    &=...\\
    &= p(s_0)\Pi^{t=k}_{t=0} \pi(a_t = y^{(t)}| s_t = (x,y^{(0:t-1)}))
\end{split}
\end{equation}
In alignment, the completed generations are of more research interest. Denoting the trajectory distribution $d^\pi(y|x)$ as the occupancy measure of completed generations conditioned on input context $x$ (i.e., final state occupancy conditioned on initial state), we have
\begin{equation}
\begin{split}
    d^\pi(y|x)=\Pi^{t=T}_{t=0} \pi(a_t = y^{(t)}| s_t = (x,y^{(0:t-1)})) = \rho^\pi(s_{{T}},a_{{T}})/p(x), \\ 
    d^\beta(y|x)=\Pi^{t=T}_{t=0} \pi_\beta(a_t = y^{(t)}| s_t = (x,y^{(0:t-1)})) = \rho^\beta(s_{{T}},a_{{T}})/p(x),
\end{split}
\end{equation}
for the current policy and behavior policy, individually. From a divergence minimization perspective, we have 
\begin{graybox}
\small
\paragraph{Divergence Minimization Perspectives of AfD (\citet{sun2024inverse}}
\begin{enumerate}[nosep,leftmargin=*]
    \item \textbf{Forward KL: SFT.} Consider the objective using the \textbf{forward KL divergence} between the demonstration and policy conditional trajectory distributions:
\begin{equation*}
\begin{split}
\label{eqn:7_true}
    \min_\pi \left[\mathrm{KL}(d^\beta(y|x)||d^\pi(y|x)) \right] = - \max_\pi \mathbb{E}_{(x,y)\sim \mathcal{D}_\mathrm{SFT}} \left[\log d^\pi(y|x) \right]
    = - \max_\pi \mathbb{E}_{(x,y^{(0:K)})\sim\mathcal{D}_\mathrm{SFT}} \left[\sum^{K}_{t=0}\log \pi(a_t|s_t) \right].
\end{split}
\end{equation*}
This corresponds to the SFT objective $\mathcal{L}_{\mathrm{SFT}} = -\max_\pi \mathbb{E}_{(s,a)\sim\mathcal{D}_\mathrm{demo}} \left[\log(\pi(a|s)) \right]$.

\item \textbf{Reverse KL: Adversarial Imitation.} Instead, minimizing the \textbf{Reverse KL divergence} leads to the following learning objective:
\begin{equation*}
\begin{split}
\label{eqn:reverse-KL_traj}
    \min_\pi [\mathrm{KL}(d^\pi(y|x)||d^\beta(y|x))] = -\max_\pi \mathbb{E}_{(x,y)\sim d^\pi}\left[ \log d^\pi(y|x) - \log d^\beta(y|x) \right],
\end{split}
\end{equation*}
Using generative adversarial methods to estimate the second term $d^\beta(y|x)$ with a parameterized discriminative model $D_\phi$, and optimizing it with
\begin{equation*}
    \max_\phi \mathbb{E}_{(y|x)\sim \mathcal{D}_{\mathrm{SFT}}}[\log D_\phi(y|x)] +  \mathbb{E}_{(y|x)\sim d^\pi}[\log (1-D_\phi(y|x))],
\end{equation*}
at convergence, the policy learning objective becomes
\begin{equation*}
\small
\label{eqn:adv_pi}
    \max_\pi \mathbb{E}_{(y|x)\sim d^\pi}\left[ r(y|x) \right],
\end{equation*}
where 
\begin{equation*}
\label{eqn:real_12}
    r(y|x) = \log D_\phi(y|x) - \log (1-D_\phi(y|x)) 
\end{equation*}
is the parameterized reward model.
\end{enumerate}
\end{graybox}
From those derivations, we see that both SFT and Reward Modeling are instantiations of divergence minimization in AfD. As the forward KL and reverse KL divergences have mass-covering and mode-seeking properties, those different objectives also lead to different model behaviors after alignment~\citep{chu2025sft}.
\begin{mdframed}[innertopmargin=0pt,leftmargin=0pt, rightmargin=0pt, innerleftmargin=10pt, innerrightmargin=10pt, skipbelow=0pt]
\textbf{\textcolor{brown}{Take-away}} \quad Beyond preference-based RLHF, alignment from demonstrations (AfD) offers a principled alternative for learning from richer supervision. By formalizing AfD through occupancy matching and divergence minimization, recent work shows that both SFT and reward modeling can be understood as special cases, paving the way for more general and theoretically grounded alignment methods.
\end{mdframed}

\subsection{Improving LLM Generation with Reward Models}
\label{sec:policy_optimization}
We discuss in this section the techniques that optimize LLM outputs using learned reward models. These methods vary in terms of whether they require model fine-tuning, whether they rely on learned value estimators, and the stage at which reward feedback is incorporated (training-time or inference-time). Table~\ref{tab:reward_optimization_methods} summarizes key approaches across this landscape.

\begin{table*}[htbp]
\centering
\small
\renewcommand{\arraystretch}{1.2}
\begin{tabular}{p{1.6cm}|p{1.0cm}|p{1.45cm}|p{3.1cm}|p{7cm}}
\toprule
\textbf{Method} & \textbf{Fine-Tune} & \textbf{Value-Estimator} & \textbf{Example Work} & \textbf{Properties} \\
\midrule
Best-of-N & No & No & \cite{stiennon2020learning,gao2023scaling,gui2024bonbon} & Simple to implement; no training needed; improves output quality. Computationally expensive in inference.\\
\hline
Iterative Tuning & Yes & No & \cite{dong2023raft,yuan2023rrhf,liu2023statistical} & Stable and effective; no RL required. Iterative training limits parallelization. \\
\hline
PPO (Classical RLHF) & Yes & Yes (GAE) & \cite{ouyang2022training,stiennon2020learning} & Well-established and widely used. Complex to train; sensitive to hyperparameters. \\
\hline
Monte-Carlo & Yes & No (MC) & \cite{li2023remax,shao2024deepseekmath,yu2025dapo} & Conceptually simple; no value network needed; strong empirical results. \\
\hline
Reward-Guided Decoding & No & No (RM) & \cite{deng2023reward,khanov2024args,liao2025reward,chen2024pad,rashid2024critical} & Allows on-the-fly control without fine-tuning. Searching can be expensive. Performance is highly dependent on fine-grained reward model fidelity. \\
\bottomrule
\end{tabular}
\caption{Overview of generation optimization methods for LLM alignment with a reward model.}
\label{tab:reward_optimization_methods}
\end{table*}

\paragraph{Best-of-N Sampling and RAFT: Filtering and Iterative Reranking}
The simplest form of reward-guided optimization is Best-of-N (BoN) sampling, where multiple candidate completions are generated, scored by a reward model, and the highest-scoring output is selected~\citep{stiennon2020learning}. This method is straightforward and requires no additional fine-tuning, but becomes computationally expensive for long-form generation or when $N$ is large.
From the performance perspective, BoN can achieve competitive performance when compared to the RL-based optimization techniques~\citep{gao2023scaling,gui2024bonbon}. And this makes BoN performance a reliable evaluation metric for reward model research~\citep{sun2024rethinking}.

A more efficient and stable alternative is to \textit{parameterize} the BoN policy through iterative supervised fine-tuning on reward-selected outputs~\citep{dong2023raft,yuan2023rrhf,liu2023statistical}. Such that in the inference time, the BoN performance can be achieved without large-scale sampling. These methods refine the model by repeatedly fine-tuning on top-ranked completions from a small candidate set, effectively incorporating reward signals without the instability and complexity of the RL-based approaches. Recent discoveries on this line of research further demonstrate its strong ability compared to state-of-the-art RL algorithms in LLM post-training~\citep{xiong2025minimalist}.

\paragraph{PPO, REINFORCE, GRPO, DAPO: From Temporal Difference to Monte-Carlo Estimation}
Among training-time methods, Proximal Policy Optimization (PPO)~\citep{schulman2017proximal} is the most widely adopted algorithm for LLM alignment~\citep{ouyang2022training,stiennon2020learning,bai2022training}. Its canonical implementation incorporates a value network and Generalized Advantage Estimation (GAE)~\citep{schulman2015high} to stabilize value propagation. However, this standard setup overlooks key differences between LLM post-training and conventional RL tasks. Unlike typical RL benchmarks~\citep{bellemare2013arcade,aitchison2023atari,tassa2018deepmind}, LLM generation receives sparse, trajectory-level feedback. For instance, correctness in mathematical reasoning is assessed only after a full solution is generated; human preference labels are typically given at the response level in chatbot alignment tasks.

This challenge is commonly referred to as the credit assignment problem in RL literature~\citep{pignatelli2023survey}, which has been tackled using techniques such as reward redistribution and decomposition~\citep{ren2021learning,arjona2019rudder}, memorization-based methods~\citep{ke2018sparse}, and attention-based mechanisms~\citep{ferret2019self}. In the LLM setting, \citet{chan2024dense} proposed a token-level redistribution scheme that leverages attention scores to assign trajectory-level rewards to individual tokens, thereby improving the stability and efficiency of PPO-based post-training.

Given the sparsity of rewards, another line of work sidesteps value estimation entirely by adopting Monte-Carlo based return estimation, such as REINFORCE~\citep{li2023remax} and GRPO~\citep{shao2024deepseekmath}, which directly optimize expected returns using trajectory-level feedback. These methods have shown strong empirical performance in tasks like mathematical reasoning and code generation. DAPO~\citep{yu2025dapo} further builds on GRPO with additional empirical insights, improving both stability and training efficiency.

\paragraph{Reward-Guided Decoding: Inference-Time Optimization without Fine-Tuning}
Unlike training-time methods that update model parameters, reward-guided decoding directly modifies the sampling procedure at inference time using a reward model to steer generation. These approaches operate by reweighting token probabilities based on token-level or trajectory-level reward feedback, offering a flexible alternative to policy model training.

Recent work has explored a range of reward-guided decoding strategies. RAD~\citep{deng2023reward} introduces a unidirectional reward model to rescore tokens during generation, improving controllability without retraining. ARGS~\citep{khanov2024args} generalizes this idea to broader alignment settings by adjusting token sampling using reward signals. PAD~\citep{chen2024pad} extends reward-guided decoding to support personalized preferences at decoding time, while RSD~\citep{liao2025reward} leverages a draft model and reward evaluation to enable efficient speculative decoding.

While promising, inference-time alignment methods often rely on trajectory-level rewards applied at the token level. \citet{rashid2024critical} highlighted such a mismatch and addressed it by training Bradley-Terry reward models on partial sequences to derive a consistent token-level policy. In more general practices, the effectiveness of such process-based reward models may vary by task and should be evaluated accordingly.

\begin{mdframed}[innertopmargin=0pt,leftmargin=0pt, rightmargin=0pt, innerleftmargin=10pt, innerrightmargin=10pt, skipbelow=0pt]
\textbf{\textcolor{brown}{Take-away}} 
LLM generation can be optimized using reward models either through training-time policy updates or inference-time decoding strategies. Besides the classical method of PPO, simpler alternatives such as iterative fine-tuning and Monte-Carlo value estimation based methods provide strong empirical performance with reduced complexity. The choice of method should consider reward sparsity, task structure, and computational constraints.
\end{mdframed}

\subsection{Risks, Challenges, and Opportunities}
\begin{figure}[h!]
    \centering
    \includegraphics[width=1.0\linewidth]{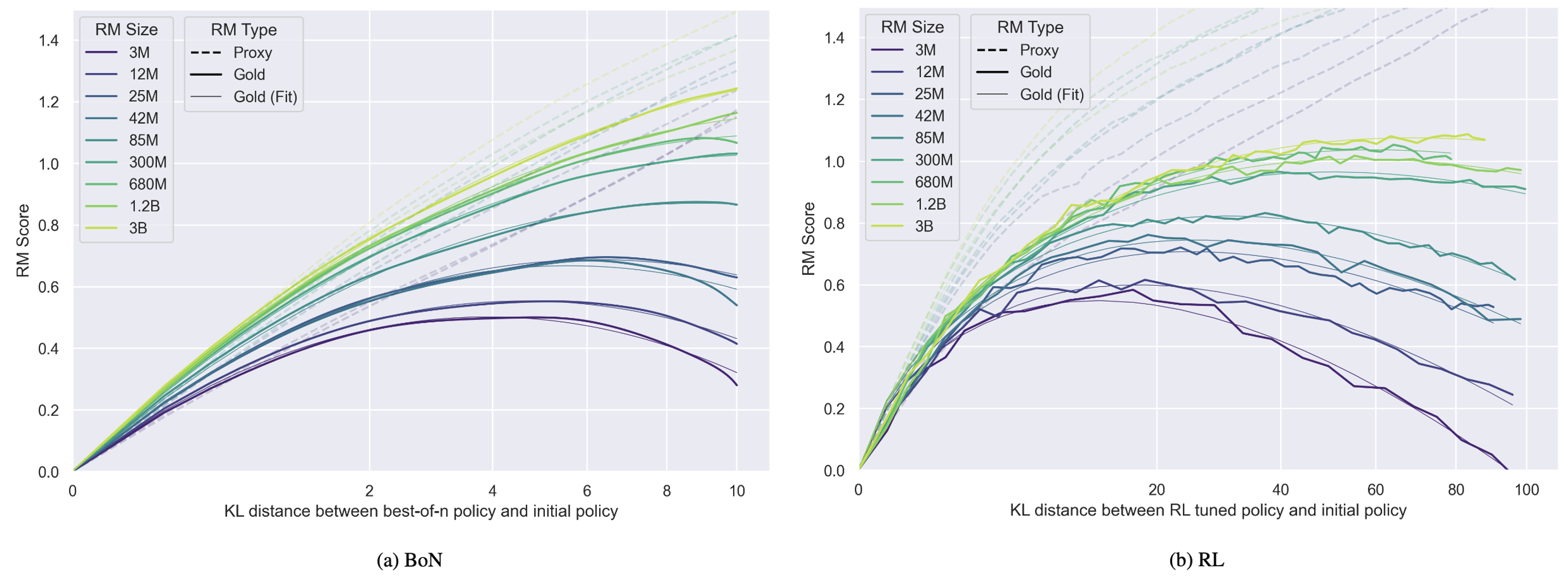}
    \vspace{-0.8cm}
    \caption{\small Reward model overoptimization (Figure 1 of \citet{gao2023scaling}). The x-axis represents the degree of optimization, measured by the KL divergence between the optimized policy and the original checkpoint. The y-axis indicates the reward score assigned by different reward models. The two panels correspond to different optimization methods: Best-of-N sampling and PPO-based training. Each curve color denotes a different reward model size. Across all settings, the gap between the solid line (score assigned by the optimized reward model) and the dashed line (score assigned by a held-out reference reward model) quantifies the degree of overoptimization—i.e., the extent to which the optimized policy exploits idiosyncrasies of the reward model rather than aligning with the intended objective.}
    \label{fig:overoptimization}
\end{figure}

\paragraph{Reward Overoptimization}
Before concluding this section, we would like to discuss some challenges and opportunities in reward modeling. Since the reward models are learned from data, it may be overfitted --- just like any data-driven machine learning models. The most well-known challenge is the reward hacking problem, or reward overoptimization~\citep{gao2023scaling}.
The key insight here is Goodhart’s Law, which states, "When a measure becomes a target, it ceases to be a good measure.”
Optimizing too much against a learned reward model will eventually hinder the true objective. As illustrated in Figure~\ref{fig:overoptimization}.

% To alleviate such a challenge, a practical approach is to include uncertainty estimation in reward modeling through ensemble methods~\citep{coste2023reward,ahmed2024scalable,zhang2024overcoming}. Another approach is to regularize the reward model learning process, with token generation as an auxiliary task in value learning~\citep{yang2024regularizing}, and the key insight behind this line of work is known as the Generative Reward Modeling(GRMs)~\citep{mahan2024generative,wang2025gram}, which leverages the natural next-token prediction abilities of LLMs to perform discriminative tasks~\citep{zhang2024generative}
% Recent work of \citet{liu2025inference} further leverage the advanced reasoning abilities of LLMs and scale inference-time compute for better reward modeling task.

To mitigate reward model overoptimization, one practical direction is to incorporate uncertainty estimation into reward modeling, often via ensemble methods~\citep{coste2023reward,ahmed2024scalable,zhang2024overcoming}. Another line of work focuses on regularizing the learning process by incorporating auxiliary objectives such as generative predictions to regularize value learning~\citep{yang2024regularizing}. This insight underpins the framework of Generative Reward Models (GRMs)~\citep{mahan2024generative,wang2025gram}, which leverages the generative capabilities of LLMs to improve reward estimation in discriminative settings~\citep{zhang2024generative}. More recently, \citet{liu2025inference} proposes to scale inference-time computation and exploit the advanced reasoning abilities of LLMs to further enhance reward modeling performance and reliability. 

Besides technical improvements, model behavior analysis can also add important insight into understanding overoptimization behaviors. 
In model evaluation, it has been discovered that users would prefer lengthy responses over concise ones, and such length bias can be captured by reward models~\citep{hu2024explaining,wang2023large,wu2023style}; hence, length-controlled evaluation has been widely adopted~\citep{dubois2024length}.
\citet{liu2024rrm} considered a causal approach to disentangle contextual artifacts and irrelevant signals, such that the robustness of reward models can be improved.

\paragraph{Data Matters: from Offline to Online Datasets}
The second challenge lies in the off-policy nature of available data. In many alignment settings, especially when leveraging open-source datasets, the responses are typically generated by outdated or mismatched models. Training reward models or optimizing policies on such off-policy data introduces distribution mismatch and can degrade performance~\citep{xiong2023gibbs}. Prior work has emphasized that data quality outweighs quantity under such conditions --- smaller, high-quality datasets often yield better results than large but stale ones~\citep{zhou2023lima,sun2024off}.

Given limited annotation budgets, online learning or active preference collection offers a more efficient alternative to static offline datasets. As discussed in Section~\ref{sec:pref_rms}, principled algorithms can help optimize annotation efforts and improve reward model quality. Future work may explore methods for converting off-policy data into usable on-policy annotations for reward modeling.

% \begin{mdframed}[innertopmargin=0pt,leftmargin=0pt, rightmargin=0pt, innerleftmargin=10pt, innerrightmargin=10pt, skipbelow=0pt]
% \textbf{\textcolor{brown}{Take-away}} 
% Summing up the challenges above, \textit{generalization} over unseen prompts, responses, and even LLM policies are the key challenge of reward modeling. Previous research on the algorithmic side has been focused on alleviating overoptimization problems, identifying reward hacking behaviors in alignment, and leveraging the advanced LLM abilities in reward modeling tasks. From the data-centric perspective, the offline data in preference learning has been shown to be a challenge. In future work, explorations on new data format, such as critique data can be further investigated.
% \end{mdframed}
\begin{mdframed}[innertopmargin=0pt,leftmargin=0pt, rightmargin=0pt, innerleftmargin=10pt, innerrightmargin=10pt, skipbelow=0pt]
\textbf{\textcolor{brown}{Take-away}} 
Across the challenges discussed above, \textit{generalization} to unseen prompts, responses, and even underlying LLM policies remains the central obstacle in reward modeling. Algorithmic advances have aimed to mitigate overoptimization, detect and analyze reward hacking, and leverage LLMs’ reasoning capabilities to improve reward modeling. On the data-centric side, the off-policy nature of preference data presents a major bottleneck. Future research may benefit from exploring diverse feedback modalities, such as critiques, and developing methods that better bridge the gap between offline supervision and on-policy learning.
\end{mdframed}

% \newpage
% \section{Infrastructure, Datasets and Benchmarks}

\bibliography{tmlr}
\bibliographystyle{tmlr}

% \appendix
% \section{Appendix}

\end{document}

%% file: math_commands.tex
\usepackage{amsmath,amsfonts,bm}

\def\eqref#1{equation~\ref{#1}}
\def\1{\bm{1}}

\DeclareMathAlphabet{\mathsfit}{\encodingdefault}{\sfdefault}{m}{sl}
\SetMathAlphabet{\mathsfit}{bold}{\encodingdefault}{\sfdefault}{bx}{n}